\documentclass{article}

\usepackage{arxiv}

\usepackage[utf8]{inputenc} 
\usepackage[T1]{fontenc}    
\usepackage{hyperref}       
\usepackage{url}            
\usepackage{booktabs}       
\usepackage{amsfonts}       
\usepackage{amsmath, amssymb}
\usepackage{nicefrac}       
\usepackage{microtype}      
\usepackage{lipsum}		
\usepackage{graphicx}
\usepackage[numbers]{natbib} 
\usepackage{doi}
\usepackage{enumitem}
\usepackage{bm}
\usepackage{multirow}

\usepackage[table]{xcolor}

\definecolor{revs}{rgb}{0, 0, 0}
\definecolor{revs-minor}{rgb}{0, 0, 0}

\title{Investigating Deep Learning Model Calibration for Classification Problems in Mechanics}

\author{{Saeed Mohammadzadeh}\\
	Division of Systems Engineering\\
	Boston University\\
	Boston, MA 02215 \\
	\texttt{saeedmhz@bu.edu} \\
        \And
	{Peerasait Prachaseree} \\
	Department of Mechanical Engineering\\
	Boston University\\
	Boston, MA 02215\\
	\texttt{pprachas@bu.edu} \\
	\And
	{Emma Lejeune} \\
	Department of Mechanical Engineering\\
	Boston University\\
	Boston, MA 02215\\
	\texttt{elejeune@bu.edu} \\
}

\date{}

\renewcommand{\shorttitle}{\textit{Investigating Deep Learning Model Calibration for Classification Problems in Mechanics}}

\hypersetup{
pdftitle={model_calibration_mechanics},
pdfsubject={deep learning model calibration for problems in mechanics},
pdfauthor={Saeed Mohammadzadeh, Peerasait Prachaseree, Emma Lejeune},
pdfkeywords={machine learning, model calibration, deep learning, open science},
}

\begin{document}
\maketitle

    \begin{abstract}
        Recently, there has been a growing interest in applying machine learning methods to problems in engineering mechanics. In particular, there has been significant interest in applying deep learning techniques to predicting the mechanical behavior of heterogeneous materials and structures. Researchers have shown that deep learning methods are able to effectively predict mechanical behavior with low error for systems ranging from engineered composites, to geometrically complex metamaterials, to heterogeneous biological tissue. However, there has been comparatively little attention paid to deep learning model calibration, i.e., the match between predicted probabilities of outcomes and the true probabilities of outcomes. In this work, we perform a comprehensive investigation into {\color{revs-minor}machine learning} model calibration across $7$ open access engineering mechanics datasets that cover three distinct types of mechanical problems. Specifically, we evaluate both model and model calibration error for multiple machine learning methods, and investigate the influence of ensemble averaging and post hoc model calibration via temperature scaling. Overall, we find that ensemble averaging of deep neural networks is both an effective and consistent tool for improving model calibration, while temperature scaling has comparatively limited benefits. Looking forward, we anticipate that this investigation will lay the foundation for future work in developing mechanics specific approaches to deep learning model calibration. 
    \end{abstract}

\section{Introduction}
\label{sec:intro} 

\begin{figure}[h]
\centering
\includegraphics[width=0.5\textwidth]{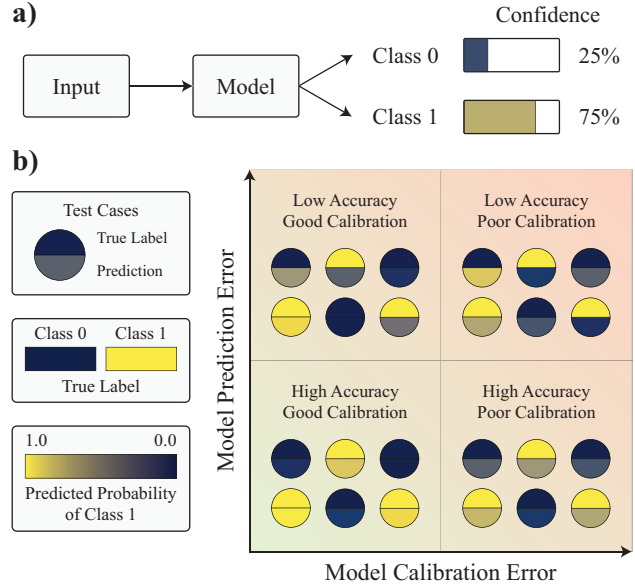}
\caption{{\color{revs}Conceptual illustration of model calibration for binary classification problems. Panel (a) depicts a standard binary classification task where the model outputs probabilities for both ``Class 0'' and ``Class 1.'' Panel (b) illustrates the combined implications of both ``model error'' and ``model calibration error'' via a toy example. In this graph, the bottom left quadrant represents high accuracy and good calibration and the top right quadrant represents low accuracy and poor calibration. In the circular test cases, the top half of each circle represents the true label, and the bottom half shows the predicted probability for class $1$. Overall, low model calibration error is achieved when the network has high confidence for correct predictions and low confidence for incorrect predictions.}}
\label{fig:intro}
\end{figure}

Over the past decade, there have been unprecedented advances in applying machine learning techniques to problems in mechanics. Researchers have used {\color{revs-minor}machine learning} approaches to enable design optimization \citep{gongora2022designing, guo2021artificial, hanakata2020forward,shin2022spiderweb, wang2020deep}, inverse analysis \citep{ardizzone2018analyzing, kakaletsis2022can, wang2019variational}, real-time predictions \citep{jin2020automated, kapteyn2021probabilistic, zandigohar2021netcut}, and multi-scale modeling \citep{alber2019integrating,karapiperis2021data, mann2022development, vlassis2020geometric, yin2022interfacing} among many other applications. There has also been a growing interest in using {\color{revs-minor}machine learning} approaches for uncertainty quantification for constitutive modeling \citep{Bayesian-EUCLID, sun2022data} and multi-fidelity surrogate modeling \citep{han2022multi, perdikaris2015multi, gander2022fast}. For these applications, which are ubiquitous in engineering design and real world decision making, it is important to investigate {\color{revs-minor}machine learning} model calibration, i.e., the match between predicted probabilities of outcomes and the true probabilities of outcomes \citep{gneiting2007strictly, guo2017calibration, minderer2021revisiting, naeini2015obtaining, niculescu2005predicting, zadrozny2002transforming}, in addition to model error. This is a particularly important consideration for deep learning models, which tend to have low model error with no associated promise of being well calibrated \citep{guo2017calibration}. In the computational mechanics community, there is a rich history of studying uncertainty quantification and model calibration \citep{arendt2012quantification, psaros2022uncertainty, wang2020perspective}. Here, our goal is to add additional context specific to deep learning based {\color{revs}\textit{classification}} problems. {\color{revs}In Fig. \ref{fig:intro}, we illustrate the concept of model calibration for binary classification problems. In Fig. \ref{fig:intro}b, we specifically highlight the relationship between model error and model calibration error.}

In the machine learning community, there is growing interest in improving model calibration without compromising predictive accuracy  \citep{guo2017calibration, minderer2021revisiting}.
For deep learning in particular, where model calibration remains poorly understood, there is a concurrent focus on \textit{evaluating} model calibration and on \textit{improving} model calibration.
Critically, there is growing attention to empirical evaluation of established model calibration metrics (e.g., expected calibration error defined in Section \ref{sec:ece}) across different deep learning model architectures \citep{guo2017calibration, minderer2021revisiting}. For example, researchers have empirically investigated the relationship between model error and model calibration error across multiple architectures on the ImageNet dataset for image classification \citep{minderer2021revisiting,deng2009imagenet}. 
Simultaneously, developing effective metrics for evaluating model calibration remains an active area of research \citep{minderer2021revisiting, naeini2015obtaining, nixon2019measuring, ovadia2019can, zhang2020mix}(discussed further in Section \ref{sec:ece_note}).

Beyond evaluating deep learning model calibration as an emergent property, there has also been significant interest in developing methods specifically for improving model calibration \citep{guo2017calibration, lakshminarayanan2017simple, platt1999probabilistic, rahaman2020uncertainty, zadrozny2002transforming, zhang2020mix}. 
Notably, interest in this topic predates the prevalent current interest in deep learning \citep{gneiting2007strictly, niculescu2005predicting}. For example, Platt Scaling, a post hoc calibration method where scores of the trained model are additionally trained through logistic regression, was initially developed for support vector machines \citep{platt1999probabilistic}. 
Broadly speaking, there are multiple post hoc calibration methods that rely on an additional held out training dataset (e.g., Platt Scaling \citep{platt1999probabilistic}, Isotonic Regression \citep{zadrozny2002transforming}) that are applicable to improving {\color{revs-minor}machine learning} model calibration. 
In the context of deep learning, deep ensembles \citep{lakshminarayanan2017simple}, temperature scaling \citep{guo2017calibration}, and combinations of deep ensembles and temperature scaling \citep{rahaman2020uncertainty, zhang2020mix} are straightforward and commonly implemented strategies for improving deep model calibration that we will investigate here.  As our ability to train large deep learning based models with low error becomes more commonplace \citep{elhassouny2019trends, guo2017calibration}, working towards better model calibration is a natural next step.

In this work, our goal is to perform a comprehensive investigation into deep learning model calibration for classification problems in mechanics. 
{\color{revs} The structure of our investigation is informed by two high level objectives. First, because deep learning model performance is \textit{dataset dependant}, it is our goal to design and implement a mechanics-specific challenge for assessing different approaches to model calibration. Namely, we want to create a multi-faceted framework to apply broad advances in machine learning to the mechanics domain. 
Second, we want to conduct a study that can be directly leveraged by others. This means that we not only want our findings to be of clear utility to others, but also that we want our framework to be directly accessible for others to build on it to assess alternative methods.
This structure is directly informed by similar investigations into deep learning methods conducted by others outside the field of mechanics \citep{kissas2022learning, minderer2021revisiting, do2020accuracy, mehrtash2020confidence}.}

{\color{revs} Following these high level goals, the} foundation for our investigation is $7$ previously published datasets that span three distinct mechanical problems, detailed in Section \ref{sec:methods_datasets}. Necessitated by the diversity in these three mechanical problems, we train distinct problem-specific deep {\color{revs-minor}learning} models on these datasets, detailed in Section \ref{sec:methods_basic}. And, across all datasets, we explore the influence of ensemble averaging, detailed in Section \ref{sec:methods_ensemble}, and temperature scaling, detailed in Section \ref{sec:methods_temp}, on model calibration. In Section \ref{sec:results}, we present the main findings from our investigation as plots of {\color{revs-minor}machine learning} model error with respect to {\color{revs-minor}machine learning} model calibration error. From this investigation, we find that ensemble averaging of deep neural networks is both an effective and consistent tool for improving model calibration for problems in mechanics, while temperature scaling has comparatively limited benefits. To our knowledge, this is the largest investigation of deep learning model calibration on open access mechanics datasets to date. It is our hope that the results of this investigation are both informative to others, and will lay the foundation for further exploration of this important topic.

\section{Methods}
\label{sec:methods} 

In this Section, we will begin by introducing the datasets used in this investigation. Please note that all datasets used in this study have been previously published by our group under Creative Commons Attribution-ShareAlike 4.0 International licenses, and are thus freely available for others to use in follow-up studies to this work. Then, in Section \ref{sec:methods_basic}, we will describe the {\color{revs-minor}machine learning} models investigated in this work. In Section \ref{sec:methods_ensemble} we will describe our implementation of ensemble averaging, and in Section \ref{sec:methods_temp} we will describe our implementation of temperature scaling. Finally, in Section \ref{sec:methods_metrics}, we will specify the error and calibration metrics used to report results in Section \ref{sec:results}. 

\subsection{Benchmark Datasets Used in this Study}
\label{sec:methods_datasets} 

In this investigation, our goal is to comprehensively evaluate model calibration on a diverse set of mechanics-based classification datasets. To this end, we will conduct our analysis on $7$ open access datasets across three types of mechanical problems. In Section \ref{sec:methods_bic}, we provide background details on the ``Buckling Instability Classification'' (BIC) dataset and sub-datasets \citep{lejeune2020buckling}, in Section \ref{sec:methods_abc}, we provide background details on the ``Asymmetric Buckling Columns'' (ABC) dataset and sub-datasets \citep{prachaseree2022asymmetric}, and in Section \ref{sec:methods_crack} we provide details on the ``Mechanical MNIST -- Crack Path'' dataset \citep{mohammadzadeh2021mechanical}. We note briefly that all datasets are derived from simulations conducted via the open source finite element analysis software FEniCS \citep{alnaes2015fenics,logg2012automated}, and the structures in the ABC dataset are generated through Gmsh \citep{geuzaine2009gmsh}. Overall, these datasets cover both a range of mechanical mechanisms (i.e., both geometric and material nonlinearity), and rely on a range of deep learning techniques (i.e., standard neural networks \citep{lejeune2021geometric}, graph neural networks \citep{prachaseree2022learning}, and convolutional networks \citep{mohammadzadeh2022predicting}). {\color{revs} Specific details for accessing each dataset and the additional background information required to recreate each dataset are provided in Section \ref{sec:additional}}.

\begin{figure}[h]
\centering
\includegraphics[width=0.5\textwidth]{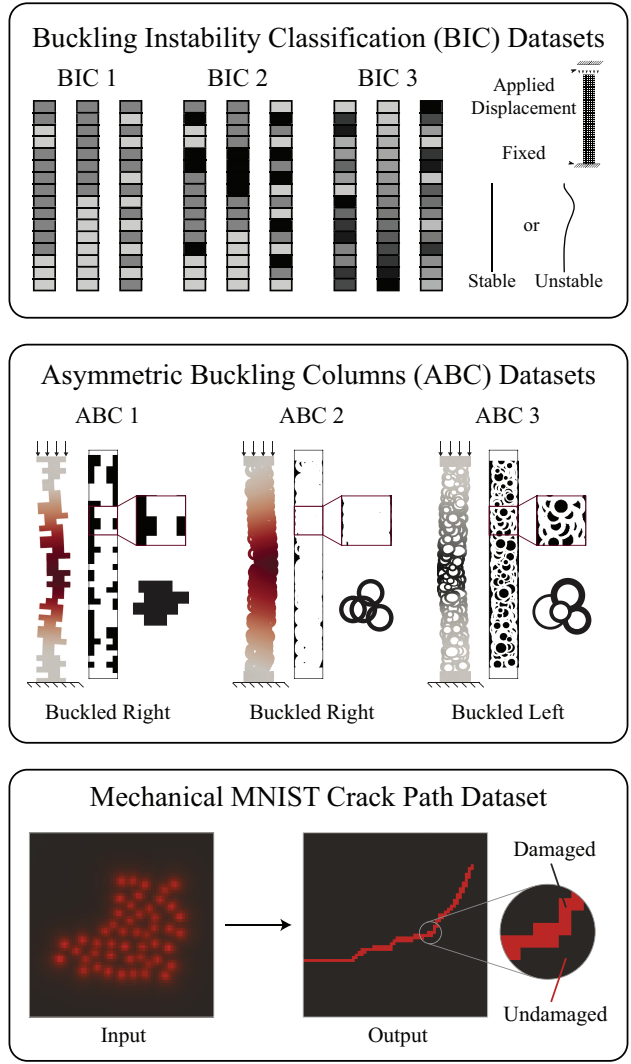}
\caption{Schematic illustration of the $7$ datasets used in this study: the BIC dataset (which contains BIC 1, BIC 2, and BIC 3) \citep{lejeune2020buckling}, the ABC dataset (which contains ABC 1, ABC 2, and ABC 3) \citep{prachaseree2022asymmetric}, and the Mechanical MNIST -- Crack Path dataset \citep{mohammadzadeh2021mechanical}.}
\label{fig:datasets}
\end{figure}

\subsubsection{Buckling Instability Classification (BIC)}
\label{sec:methods_bic}

The BIC dataset was disseminated in conjunction with our previous publication exploring multiple straightforward approaches to classification problems in mechanics \citep{lejeune2021geometric}. The BIC dataset contains three sub-datasets: ``BIC 1'', ``BIC 2'', and ``BIC 3'' that differ only in their input parameter distribution. In all cases, the input is the material property distribution of a heterogeneous column that is then subject to a fixed level of applied compressive displacement. The output then corresponds to class ``Stable'' or ``Unstable'' based on the results of a Finite Element Analysis (FEA) simulation. For all sub-datasets, the input property distribution of each sample is represented by a $16 \times 1$ vector. For BIC 1, there are two possible discrete modulus $E$ values: $E=1$, and $E=4$. For BIC 2, there are three possible discrete values: $E=1$, $E=4$, and $E=7$. For BIC 3, the modulus varies continuously to three degrees of precision in the range $E=[1,8]$. Further details regarding data curation and our FEA implementation are available in our previous publication \citep{lejeune2021geometric}. For the work presented in this manuscript, we used $10,000$ samples for {\color{revs-minor}machine learning} model training and $1,000$ samples for post hoc model calibration across all three sub-datasets, and $6,553$ samples for BIC 1 and $10,000$ samples for BIC 2 and BIC 3 for model testing. Briefly, we note that in Section \ref{sec:results_1}, we also specifically investigate training set sizes of $200$, $500$, $1000$, $2000$, $5000$, $10000$ samples. For the BIC 1 dataset, the ratio between ``Stable'' and ``Unstable'' samples is $0.28$, for BIC 2 it is $0.40$, and for BIC 3 it is $0.26$.

\subsubsection{Asymmetric Buckling Columns (ABC)}
\label{sec:methods_abc}
As a follow up to the BIC dataset, we introduced the ABC dataset \citep{prachaseree2022asymmetric} in our previous work in conjunction with an exploration of geometric deep learning for mechanics-specific classification problems \citep{prachaseree2022learning}. Similar to BIC, the ABC dataset contains three subdatasets, where each subdataset corresponds to a different algorithm for generating the geometry of the input domain. For ABC 1, columns are generated by vertically stacking rectangular blocks of randomly varying widths. For ABC 2, columns are generated by randomly overlaying rings of identical inner and outer radii. For ABC 3, columns are generated by overlaying and \textit{trimming} rings of varying inner and outer radii (i.e., varying size and thickness). For all sub-datasets, the columns are subjected to fixed-fixed boundary conditions and are compressed until the onset of buckling. Each input geometry in the ABC dataset is then classified as buckling ``left'' or ``right.'' Further details of data curation, and FEA implementation are available in our previous publication \citep{prachaseree2022learning}. For the work presented in this manuscript, we used $20,000$ samples for {\color{revs-minor}machine learning} model training, $1,000$ samples for post hoc model calibration, and $2,500$ samples for model testing for each of the three sub-datasets. For all three ABC sub-datasets, the classes are balanced.

\subsubsection{Mechanical MNIST -- Crack Path}
\label{sec:methods_crack}

The Mechanical MNIST dataset collection \citep{lejeune2020EML, MNIST_CH, MNIST_DS} is a collection of benchmark datasets initially conceptualized as mechanics-relevant drop-in replacements for the popular MNIST dataset \citep{mnist}. For the datasets in the Mechanical MNIST collection, input bitmaps dictate heterogeneous material properties, and outputs are defined as curated results from FEA simulations. The ``Mechanical MNIST -- Crack Path'' dataset is an example from the collection where the input bitmap distribution is a heterogeneous pattern of inclusions derived from the Fashion MNIST dataset \citep{fashion-mnist}, and the main output is a damage field predicted by a linear elastic phase-field fracture simulation \citep{phase-field-review,phase-field-unified}. For this manuscript, we will focus exclusively on the $64\times64$ input bitmap and a downsampled $64\times64$ output crack path. Notably, each pixel in this downsampled crack path is in either the ``damaged'' (true) class or the ``undamaged'' (false) class, thus conceptualizing the Mechanical MNIST -- Crack Path dataset as a binary classification problem similar to the BIC and ABC datasets described previously. Further details of data curation and our FEA implementation are available in our previous publication \citep{mohammadzadeh2022predicting}. {\color{revs}In the original dataset there are $60000$ samples in the training set and $10000$ samples in the test set.} For the work in this manuscript, we used {\color{revs}the first} $10,000$ samples {\color{revs}from the training set} for {\color{revs-minor}machine learning} model training, {\color{revs}the next} $1,000$ samples {\color{revs}from the training set} for post hoc model calibration, and $10,000$ samples {\color{revs}from the test set} for model testing. For this dataset, classes are heavily imbalanced, where the ``damaged'' class corresponds to $2.88 \%$ of pixels. 

\subsection{Machine Learning Models Investigated}
\label{sec:methods_basic} 

The main focus of this work is on \textit{deep learning} model calibration. However, in Section \ref{sec:res_eval}, we provide baseline comparisons to Gaussian Process Classification and Support Vector Classification to add additional context to our results. Here we briefly summarize these methods along with the Neural Network based approaches used for prediction. 

\subsubsection{Gaussian Process Classification}
\label{sec:gpc}
Gaussian Processes are commonly used in machine learning literature for both classification and regression tasks when uncertainty quantification is critical \citep{bartok2022improved}. The Gaussian Process Classification \citep{williams2006gaussian}, which we use in this work, is a generalization of the linear logistic regression model where the linear latent function is replaced by a Gaussian Process. 
To train a Gaussian Process Classification in the context of a machine learning problem, the user must define a kernel function that will determine the form of the covariance matrix. For further details on Gaussian Process methods in machine learning, we refer the reader to the literature \citep{williams2006gaussian}. In this work, we used scikit-learn \citep{scikit-learn} to train Gaussian Process Classification with a Radial Basis Function kernel \citep{duvenaud2014kernel} on the BIC 1, BIC 2, and BIC 3 datasets. The performance of Gaussian Process Classification on these data is shown in Section \ref{sec:res_eval}. The code for this model is included in the GitHub Repository for this publication.

\subsubsection{Support Vector Classification}
\label{sec:svc}

Support Vector Machines are a commonly used machine learning algorithm for classification problems \citep{hearst1998support}. The Support Vector Machine was initially developed for binary classification problems and later on extended to deal with multi-class classification \citep{weston1998multi} and regression tasks \citep{drucker1996support}. 
Here, we will focus on ``Support Vector Classification'' for binary classification. In brief, the general idea behind Support Vector Classification is to transform the data in a high dimensional space and identify a hyperplane that most accurately separates the classes. Similar to Gaussian Process Classification, choice of kernel function impacts Support Vector Classification performance in the context of machine learning. 
In this work, we use scikit-learn \citep{scikit-learn} to train Support Vector Classifications with a Radial Basis Function kernel and no additional regularization on the BIC 1, BIC 2, and BIC 3 datasets. The performance of Support Vector Classifications on these data is shown in Section \ref{sec:res_eval}. Critically, we note that scikit-learn uses Platt scaling, described in \citep{platt1999probabilistic}, coupled with five-fold cross-validation to obtain probabilistic outputs from otherwise non-probabilistic Support Vector Classification scores. The code for this model is included in the GitHub Repository for this publication.

\subsubsection{Fully Connected Neural Network}
\label{sec:fnn}

Fully Connected Neural Networks are a well-established method for both regression and classification tasks. These networks commonly consist of an input layer, an output layer, and a series of fully connected hidden layers. Each layer applies a linear transformation followed by a non-linear activation function such as Rectified Linear Units (ReLU) \citep{agarap2018deep} on its input vector. These layers are designed to eventually transform a given input vector into an output vector of the desired size. For the BIC datasets, we use Fully Connected Neural Networks (simply referred to as neural networks) with a $16$ node input layer, three $200$ node hidden layers, and a $2$ node output layer {\color{revs}(see Appendix \ref{sec:apx_neural_nets})}. The $2$ output nodes are the logits indicating to which class ``stable'' or ``unstable'' a sample belongs. To reduce overfitting during training, we add batch normalization \citep{ioffe2015batch} before applying each ReLU activation function and use dropout \citep{srivastava2014dropout} with the rate of $0.5$ before the second and third hidden layers and output layer. We use the PyTorch library \citep{paszke2019pytorch} for our implementation, and train each network for $50$ epochs using the Adam optimizer \citep{kingma2014adam} with an initial learning rate of $0.001$, which is dropped to $0.0005$ after $25$ epochs. We train this model on the BIC 1, BIC 2, and BIC 3 datasets, and present the results in both Section \ref{sec:res_eval} and Section \ref{sec:res_improve}. The code for this model is included in the GitHub Repository for this publication.

\subsubsection{Graph Neural Network}
\label{sec:gnn}
By design, the ABC datasets contain complex geometries that are attractive to represent as spatial graphs rather than as ``image-like'' arrays. In our previous work, we identified a set of best performing models and data representation approaches for each ABC subdataset that we directly build on for this work \citep{prachaseree2022learning}. In brief, each ABC input geometry is first represented as a spatial graph. In our previous work, we identified a high performing strategy for spatial graph representation where spatial graphs are constructed via discretizing the structure into nodes and then performing a ball query to form edges. Based on our prior investigation, ABC 1 had a ``medium'' node density ($\approx 306$ nodes per structure) with a ball radius of $40\%$ of the column width, and ABC 2 and ABC 3 have a ``dense'' node density ($\approx 566$ and $\approx 768$ nodes per structure respectively) with a ball radius of $30\%$ of the column width \citep{prachaseree2022learning}. 
Given this spatial graph representation, we used PointNet++ layers \citep{pointnet} as spatial graph convolution layers coupled with batch normalization \citep{ioffe2015batch}, followed by skip connections and a linear classifier to construct our machine learning model {\color{revs}(see Appendix \ref{sec:apx_neural_nets})}. To improve model performance, we also augment our dataset by flipping the columns along the $x$ axis, $y$ axis, and both axes while changing labels as needed. All models are implemented with the Pytorch Geometric library \citep{PyG} and trained using the Adam optimizer \citep{kingma2014adam} for $50$ epochs. We present the results from this model on the ABC datasets in Section \ref{sec:res_improve}. The code for the model is published on GitHub alongside our previous publication.

\subsubsection{UNet Neural Network}
\label{sec:unet}

To complement the models described in Section \ref{sec:gpc} - \ref{sec:gnn} which are trained to predict a single quantity of interest, we train a deep neural network on the Mechanical MNIST -- Crack Path dataset that is designed to predict full-field quantities of interest, specifically the whole domain damage field. In our previous work \citep{mohammadzadeh2022predicting}, we used a modified version of the UNet model \citep{ronneberger2015u}, the MultiRes-WNet, combined with a convolutional autoencoder for an end-to-end prediction of $256\times256$ images of the damage field from $64\times64$ material distribution input images. Here, we regenerated lower resolution output damage fields directly from our FEA results as $64\times64$ arrays and used a standard UNet with three downsampling and upsampling steps \citep{siddique2021u}. The outputs of the model are logits in the form of two-channel images that can be transformed into probabilities by applying a softmax function to each pixel {\color{revs}(see Appendix \ref{sec:apx_neural_nets})}. We briefly note that we trained the network by minimizing the Dice-loss \citep{jadon2020survey}. We use the PyTorch library \citep{paszke2019pytorch} for the UNet model implementation, and train each network for $50$ epochs using the Adam Optimizer \citep{kingma2014adam}. We present the results from this model on the Mechanical MNIST -- Crack Path dataset in Section \ref{sec:res_improve}. The code for this model is included in the GitHub Repository for this publication.

\subsection{Ensemble Methods}
\label{sec:methods_ensemble} 

For the Neural Network approaches introduced in Section \ref{sec:fnn}-\ref{sec:unet}, the behavior of each trained neural network will vary based on the random weight initialization. Thus, it is possible to train multiple neural networks and subsequently combine them into an ensemble \citep{ciregan2012multi,lakshminarayanan2017simple}. Here, we take a straightforward approach and individually train $10$ models with different initialization seeds before aggregating the predictions using soft voting, also referred to as unweighted model averaging \citep{lakshminarayanan2017simple}. In soft voting, the predicted probability for each class is averaged over all models and the label with the highest probability then becomes the final class prediction. The goal of ensemble averaging is to increase the overall prediction accuracy. Additionally, if the neural networks are trained with proper scoring rules like cross entropy, ensemble averaging may also lead to averaged probabilities that are well calibrated \citep{lakshminarayanan2017simple}. One major goal of this work is to critically evaluate the efficacy of neural network ensemble averaging for deep learning approaches to classification problems in mechanics. 

\subsection{Post Hoc Calibration via Temperature Scaling}
\label{sec:methods_temp} 
Post hoc calibration of neural networks using a held-out calibration dataset is a popular approach with both parametric (e.g., Platt scaling \citep{platt1999probabilistic}, temperature scaling, matrix scaling \citep{guo2017calibration}) and non-parametric (e.g., Bayesian Binning \citep{naeini2015obtaining}, isotonic regression \citep{zadrozny2002transforming}) implementations. Motivated by its popularity in the literature, we choose temperature scaling as a standard post hoc calibration technique. Specifically, given a trained classifier, we divide the logits vector $\mathbf{z}$ by a single variable $T$ called the temperature. The optimal temperature is obtained by minimizing the Negative Log Likelihood (NLL) on the held out calibration set. The NLL is written as:
\begin{equation}
    \begin{aligned}
        \min_{T} \quad & \sum_{i=1}^{N_c} \text{NLL} \bigg(\sigma({\mathbf{z}_i \, / \, T}) \, , \, y_i\bigg)\\
        \textrm{s.t.} \quad & T > 0\\
    \label{temp_scaling}
    \end{aligned}
\end{equation}
where $\sigma(\mathbf{x})$ is the softmax function, $y$ is the true labels, $N_c$ is the number of sample points in the calibration set, and $\mathbf{z}$ is the previously defined logits vector. Notably, temperature scaling can be applied either before or after ensemble averaging \citep{rahaman2020uncertainty}.

In Section \ref{sec:res_improve}, we report the results of applying post hoc calibration methods on our datasets. For clarity, the methods investigated are defined as follows:
\begin{itemize}
\item \underline{Method \textbf{I}:} Individual neural network \textit{without} post hoc calibration. 
\item \underline{Method \textbf{I-C}:} Individual neural network \textit{with} post hoc calibration via temperature scaling. 
\item \underline{Method \textbf{E-M1}:} Ensemble neural network \textit{without} post hoc calibration. 
\item \underline{Method \textbf{E-M2}:} Ensemble neural network \textit{with} post hoc calibration via temperature scaling applied \textit{before} ensemble averaging. 
\item \underline{Method \textbf{E-M3}:} Ensemble neural network \textit{with} post hoc calibration via temperature scaling applied \textit{after} ensemble averaging. 
\end{itemize}
Fig. \ref{fig:calibration} in Section \ref{sec:res_improve} and Fig. \ref{fig:reliability_2} in Appendix \ref{sec:apx_1} directly reference these definitions. Our code for implementing temperature scaling is included in the GitHub Repository for this publication.

\subsection{Error and Calibration Metrics Reported in this Investigation}
\label{sec:methods_metrics} 
In this work, all supervised learning tasks are binary classification problems. For the datasets without severely imbalanced class labels (i.e. BIC and ABC) we report classification accuracy as our error metric, while for the severely imbalanced class label (i.e. damaged pixels in the Mechanical MNIST Crack Path dataset) we report the $F_1$ score as our error metric. To measure calibration, we report the Expected Calibration Error (ECE). In all cases, we report these metrics on our held out test datasets. Details on metrics for model error and model calibration error are as follows. 

\subsubsection{Classification Error Definition for BIC and ABC}
We evaluate model performance for the BIC and ABC datasets via traditional classification error. Specifically, we define classification error as the fraction of wrong predicted labels with respect to the total number of labels. Mathematically, this is written as:

\begin{equation}
    \text{Error}(y, \, \hat{y}) = \frac{1}{N} \sum_{i=1}^N \mathbf{1}(\hat{y}_i \neq y_i) 
    \label{class_error}
\end{equation}

where $N$ is the number of labels to evaluate, $y$ and $\hat{y}$ are the true and predicted labels respectively, and $\mathbf{1}(\hat{y} \neq y)$ represents the 0-1 loss function \citep{shalev2014understanding}.

\subsubsection{Classification Error Definition for Mechanical MNIST Crack Path}

Following our previous work, the damage field for each sample in the Mechanical MNIST dataset is treated as a binary matrix (image) where $1$ represents a damaged sub-region (pixel) and $0$ represents an undamaged sub-region (pixels) \citep{mohammadzadeh2022predicting}. Because the damaged region represents a crack path, the relative prevalence of damaged pixels is small ($2.88\%$) leading to severe class imbalance. As such, the classification error as defined in eqn. \ref{class_error} will automatically appear as a small value due to the high percentage of true negative pixels in each prediction. To better evaluate model performance with these imbalanced labels, we report the error as the S{\o}rensen-Dice index, often referred to as the $F_1$ score, defined as:

\begin{equation}
    F_1 = \frac{2 \, \text{{\color{revs-minor}True Positive}}}{2 \, \text{{\color{revs-minor}True Positive}} + \text{{\color{revs-minor}False Positive}}+ \text{{\color{revs-minor}False Negative}}}
    \label{F_1}
\end{equation}

where $\text{{\color{revs-minor}True Positive}}$, $\text{{\color{revs-minor}False Positive}}$, and $\text{{\color{revs-minor}False Negative}}$ denote the number of {\color{revs-minor}correctly predicted damaged pixels}, {\color{revs-minor}incorrectly predicted undamaged pixels}, and {\color{revs-minor}incorrectly predicted damaged pixels} respectively. We note that that $F_1$ score defined in eqn. \ref{F_1} is not influenced by the number of true negative pixels {\color{revs-minor}(correctly predicted undamaged pixels)}, making it an easier-to-interpret metric of model predictive performance for the Mechanical MNIST -- Crack Path dataset. In Fig. \ref{fig:calibration}, where we report $F_1$ score, we plot error as $1 - F_1$ on the y axis to maintain visual consistency.

\begin{figure}[h]
\centering
\includegraphics[width=0.5\textwidth]{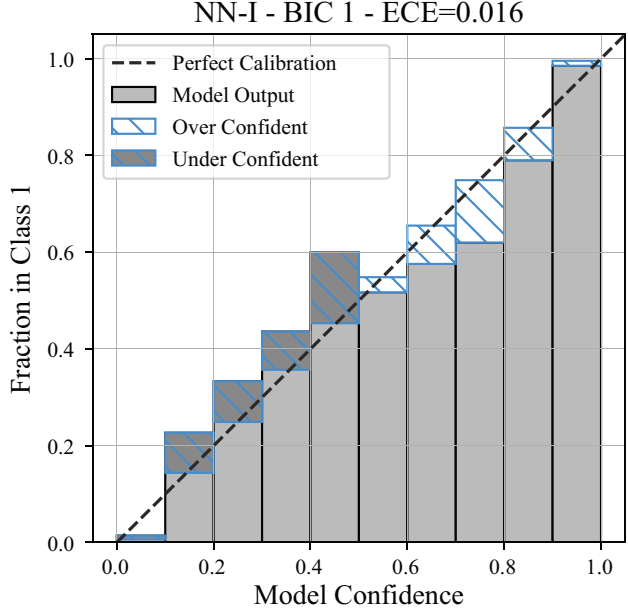}
\caption{Representative reliability diagram for an individual neural network trained on the BIC 3 dataset. To promote completeness while preserving data visualization clarity, we report additional reliability diagrams in Appendix \ref{sec:apx_1} as a supplement to Section \ref{sec:results}.}
\label{fig:reliability}
\end{figure}

\subsubsection{Expected Calibration Error (ECE)}
\label{sec:ece}

In this work, we use the Expected Calibration Error (ECE) to evaluate model calibration. As stated in Section \ref{sec:intro}, calibration refers to the match between predicted probabilities of outcomes and the true probabilities of outcomes. For example, when a model is perfectly calibrated, $100$ predictions with $80\%$ confidence should be correct $80 / 100$ times. While there are many potential metrics used to evaluate calibration \citep{guo2017calibration, minderer2021revisiting, naeini2015obtaining, niculescu2005predicting, nixon2019measuring, ovadia2019can, zhang2020mix}, the ECE is one of the most prevalent in the literature and one of the most interpretable. Namely, the ECE is connected to the reliability diagram, a common approach to visualizing model calibration illustrated in Fig. \ref{fig:reliability} \citep{guo2017calibration, naeini2015obtaining, niculescu2005predicting}. Reliability diagrams are constructed in two steps. First, sample prediction confidences (i.e., confidence that a given sample is in a chosen class) are binned. In this work, we use $10$ equally spaced bins to construct all reliability diagrams. Then, the average bin confidence is compared to the true fraction of samples with the chosen class. The ECE is then computed as the weighted average of the gap between perfect calibration and model confidence within each bin. Mathematically, this is defined as:
\begin{equation}
    \text{ECE} = \sum_{i=1}^B \frac{n_i}{N} \, | \, F_i-C_i \,|
    \label{eqn:ece}
\end{equation}
where $B$ is the number of bins, $n_i$ is the number of samples in each bin, $N$ is the total number of samples, $F_i$ is the frequency of the chosen class in the bin, and $C_i$ is the average confidence that the sample is in the chosen class in the bin. Following this definition, a lower ECE corresponds to a better calibrated model, and a model with ECE$=0$ corresponds to a perfectly calibrated model. As indicated in Fig. \ref{fig:reliability}, the reliability curve bins in a perfectly calibrated model will follow the diagonal $y=x$.

\subsubsection{Note on the Limitations of Expected Calibration Error (ECE)}
\label{sec:ece_note}


As stated previously, we report ECE as our main model calibration metric because it is both prevalent in the literature and relatively interpretable. However, it is not without limitations. For example, ECE is known to be sensitive to the selection of binning scheme \citep{nixon2019measuring, ovadia2019can, zhang2020mix}. More specifically, ECE values can depend on the bin size as well as the number of samples in each bin. Multiple modifications to the definition of ECE defined in Section \ref{sec:ece} such as adaptive binning schemes \citep{nixon2019measuring}, using the $\ell_2$ norm instead of the $\ell_1$ norm to compute the ECE \citep{minderer2021revisiting, nixon2019measuring}, and kernel-density based methods \citep{zhang2020mix}, have been proposed in the literature. Aside from the ECE, other methods that are related to reliability diagrams like the Maximum Calibration Error \citep{naeini2015obtaining} have been proposed. Alternatively, proper scoring rules \citep{gneiting2007strictly} with roots in statistical analysis like the Negative Log-Likelihood (NLL) and Brier Score \citep{minderer2021revisiting, ovadia2019can} have been used to evaluate model calibration. However, it is not clear if these proposed methods are significantly better than the standard method for computing ECE, and these new metrics potentially lose some of the clear relationship to the interpretable reliability diagram. Looking forward, we anticipate that the framework we establish in this paper could also be used to investigate the behavior of these alternative metrics. However, this is beyond the scope of our current work.

\begin{figure}[h]
\centering
\includegraphics[width=0.5\textwidth]{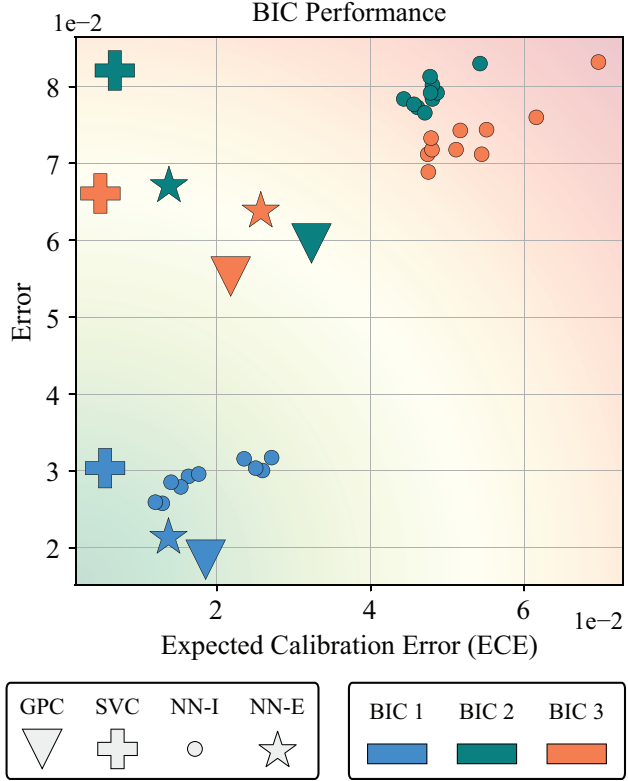}
\caption{Error vs ECE plots for Gaussian Process Classification (GPC), Support Vector Classification (SVC), 10 individual neural networks (NN-I), and an ensemble of 10 neural networks (NN-E) trained on the BIC 1, BIC 2, and BIC 3 datasets. Lower ECE and error indicate better performance (bottom left corner, represented by the green background gradient). Figure \ref{fig:reliability_1} in Appendix \ref{sec:apx_1} contains reliability diagrams that supplement these results.}
\label{fig:bic-err-vs-ece}
\end{figure}

\section{Results and Discussion}
\label{sec:results} 

In Section \ref{sec:methods}, we introduced $7$ datasets (BIC 1, BIC 2, BIC 3, ABC 1, ABC 2, ABC 3, and Mechanical MNIST -- Crack Path), described multiple {\color{revs-minor}machine learning} models for making predictions with these datasets, procedures for ensemble averaging and temperature scaling, and metrics for evaluating model error and calibration. Here, we will begin in Section \ref{sec:res_eval} by comparatively evaluating different {\color{revs-minor}machine learning} methods on the BIC datasets. Then, in Section \ref{sec:res_improve}, we will investigate multiple strategies for improving model calibration across the BIC, ABC, and Mechanical MNIST -- Crack Path datasets. {\color{revs}Throughout this Section, we present the results of our study following the format introduced in Fig. \ref{fig:inro} where each model is represented as a marker on a prediction error vs. model calibration error axis.}

\subsection{Evaluating Model Calibration}
\label{sec:res_eval}

As discussed in Section \ref{sec:intro}, large deep neural networks are prone to being poorly calibrated \citep{guo2017calibration}. In response to this claim, our first goal is to evaluate this statement in the context of our mechanics-specific datasets and establish baseline results that will contextualize future endeavors for improving model calibration.

\subsubsection{Ensemble Averaging Improves the Performance of Deep Neural Networks Across the BIC 1, BIC 2, and BIC 3 Datasets}
\label{sec:results_1}

In Fig. \ref{fig:bic-err-vs-ece}, we plot model error vs. model expected calibration error (ECE) for multiple {\color{revs-minor}machine learning} models trained on the BIC 1, BIC 2, and BIC 3 datasets. Note that each marker corresponds to the test performance of a trained {\color{revs-minor}machine learning} model. To add context to our investigation of deep learning model calibration, we show the performance of Gaussian Process Classification and Support Vector Classification with Platt scaling alongside the performance of neural networks. Consistent with results from our prior publication \citep{lejeune2021geometric}, model performance across different BIC sub-datasets varies {\color{revs-minor} due to the different input parameter space for each BIC dataset. A machine learning problem with larger input parameter space (i.e. BIC 3) will typically result in a higher error than a machine learning problem with smaller input parameter space (i.e. BIC 1).} In Fig. \ref{fig:bic-err-vs-ece}, we see that individual neural networks have a large ECE range, and tend to be poorly calibrated in comparison to the Gaussian Process Classification and Support Vector Classification models. However, ensemble neural networks have both lower error and lower ECE, and perform similarly to their Gaussian Process Classification and Support Vector Classification counterparts. This context is important because it justifies the choice of ensemble neural networks as an approach to designing well calibrated deep learning based model frameworks. In Appendix \ref{sec:apx_1}, Fig. \ref{fig:reliability_1}, we show supplementary reliability diagrams that further support these results.

\begin{figure}[h]
\centering
\includegraphics[width=0.95\textwidth]{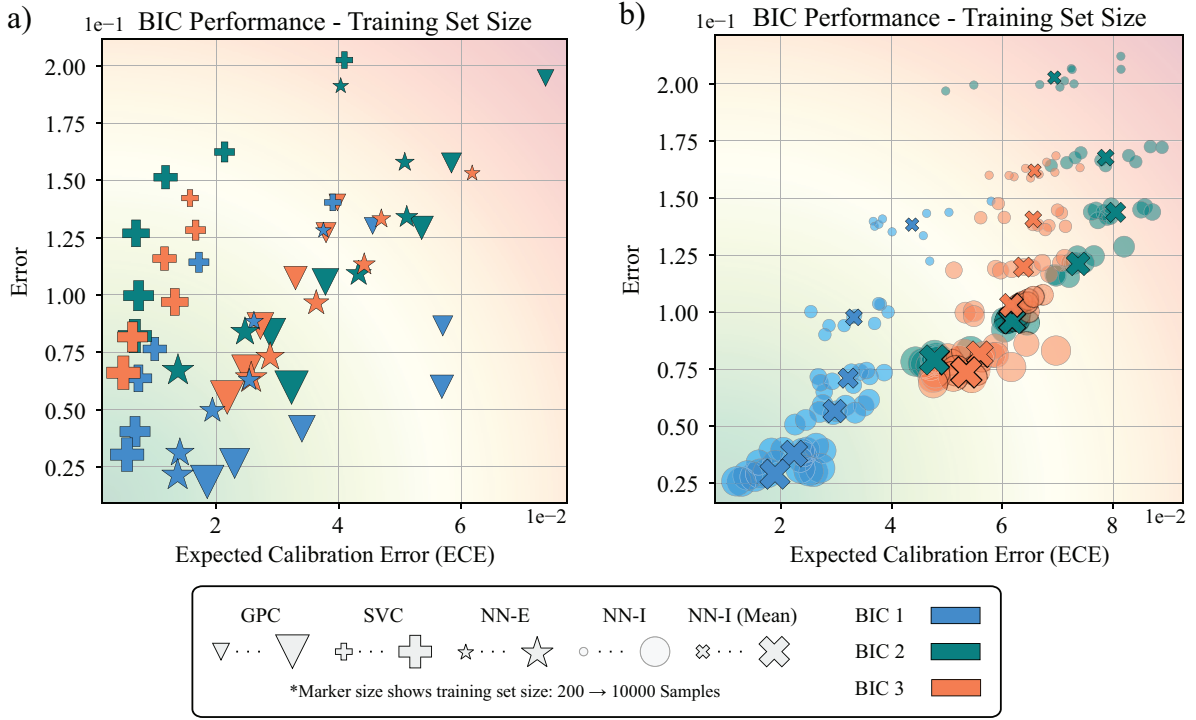}
\caption{Error vs ECE plots with varied training set sizes: a) Gaussian Process Classification (GPC), Support Vector Classification (SVC), and an ensemble of $10$ neural networks (NN-E); b) $10$ individual neural networks (NN-I) and mean values of these $10$ networks (NN-I (Mean)). All models are trained on multiple subsets of the BIC $1$, BIC $2$, and BIC $3$ datasets with varying training set sizes ($200$, $500$, $1,000$, $2,000$, $5,000$, and $10,000$ samples). Lower ECE and error indicate better performance (bottom left corner, represented by the green background gradient). The Pearson correlation coefficients for error and ECE on BIC $1$, BIC $2$, and BIC $3$, respectively, are as follows. SVC: ($0.899$, $0.887$, $0.917$); GPC: ($0.691$, $0.967$, $0.995$); NN-E: ($0.981$, $0.731$, $0.983$); NN-I: ($0.850$, $0.612$, $0.659$); NN-I (Mean): ($0.969$, $0.656$, $0.936$).}
\label{fig:training-size}
\end{figure}

\subsubsection{Increasing the Training Set Size Leads to Both Lower Error and Lower Expected Calibration Error (ECE) in Ensemble Neural Networks Across BIC 1, BIC 2, and BIC 3}
\label{sec:results_1}

In Fig. \ref{fig:training-size}, we plot model error vs. model expected calibration error (ECE) for multiple model types with different training set sizes trained on the BIC 1, BIC 2, and BIC 3 datasets. In Fig. \ref{fig:training-size}, each marker corresponds to the test performance of a trained {\color{revs-minor}machine learning} model, and marker size indicates the size of the training dataset.  Overall, Fig. \ref{fig:training-size} illustrates two important results. First, it is clear from the distribution of results across all {\color{revs-minor}machine learning} models and datasets that model error and ECE are at most weakly correlated (the overall correlation coefficient across all models is $0.516$). Namely, low model error does not necessarily indicate low ECE, and high model error does not necessarily indicate high ECE. Second, if we examine individual model types (i.e., Gaussian Process Classification, Support Vector Classification, or ensemble neural networks) and individual datasets (i.e., BIC 1, BIC 2, BIC 3) increasing the training set size consistently lowers both model error and model ECE. For example, for ensemble neural networks the correlation coefficients relating error and ECE are $0.981$, $0.731$, and $0.983$ for BIC 1, BIC 2, and BIC 3 respectively. Critically, for these examples, increasing the training set size improves both model error and model calibration which is an important observation because it is an actionable strategy for improving both dimensions of performance. 

\subsection{Improving Model Calibration}
\label{sec:res_improve}

In Sections \ref{sec:methods_ensemble} and \ref{sec:methods_temp}, we introduced two strategies for improving model calibration. The first, ensemble averaging, relies on training multiple neural networks on the same data using different random weight initialization. The second, post hoc calibration via temperature scaling, relies on reserving additional data for calibration. In this Section, we will compare the performance of individual neural networks (I), individual neural networks with post hoc calibrated via temperature scaling (I-C), ensemble averaging (E-M1) and two different methods for ensemble averaging combined with post hoc calibrated via temperature scaling (E-M2, and E-M3, see Section \ref{sec:methods_temp} for definitions). {\color{revs-minor} Similar to machine learning literature,} we will investigate all approaches across the BIC, ABC, and Mechanical MNIST -- Crack Path datasets in order to identify outcomes that are potentially consistent across diverse types of mechanical data. {\color{revs} We note that, as expected, Fig. \ref{fig:training-size} shows that error generally decreases as the training set size increases, but eventually reaches a point of ``diminishing returns'', where additional training data only marginally improves the accuracy. This result is consistent with the standard observation in machine learning and highlights the importance of selecting an appropriate training set size.}

\begin{figure}[h]
\centering\includegraphics[width=0.995\textwidth,height=\textheight,keepaspectratio]{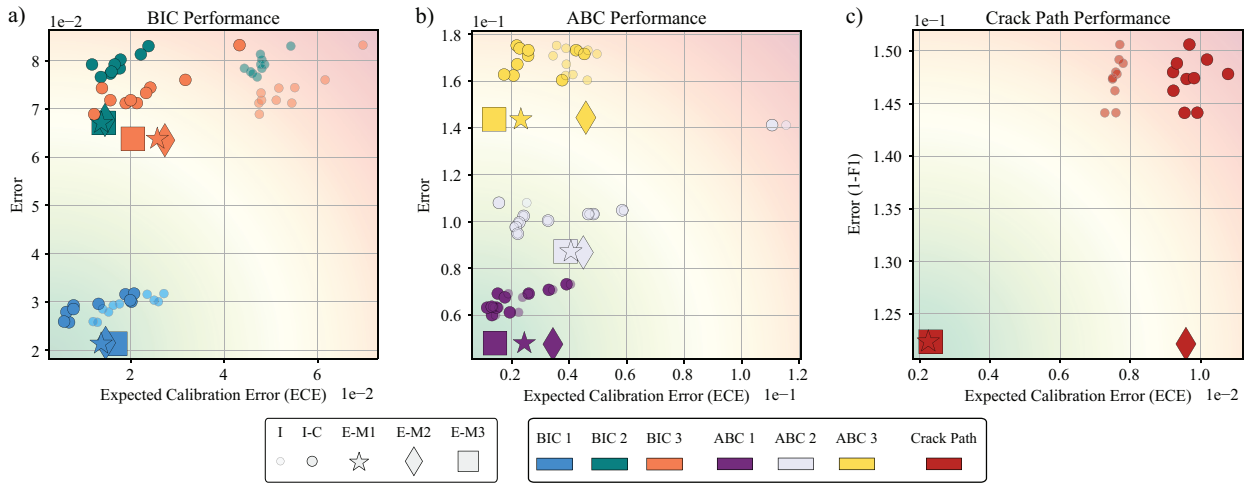}
\caption{Visualization of the influence of ensemble averaging and post hoc model calibration on error and ECE for (a) the BIC datasets, (b) the ABC datasets, and (c) the Mechanical MNIST Crack Path dataset. Color indicates the (sub)dataset, and marker style specifies the ensemble averaging and calibration approach, see Section \ref{sec:methods_temp}. Lower ECE and error indicate better performance (bottom left corner, represented by the green background gradient). Fig. \ref{fig:reliability_2} in Appendix \ref{sec:apx_1} contains reliability diagrams that supplement these results.}
\label{fig:calibration}
\end{figure}

\subsubsection{Ensemble Averaging Improves Deep Learning Model Calibration Across All Datasets}

Consistent with the results presented in Section \ref{sec:results_1}, we find that ensemble averaging without post hoc calibration (E-M1) improves error and ECE across all datasets. We note briefly that in Fig. \ref{fig:calibration}a, which represents the BIC datasets, the I and E-M1 datapoints are repeated from Fig. \ref{fig:bic-err-vs-ece}, and in Fig. \ref{fig:calibration}b, which represents the ABC datasets, the I and E-M1 datapoints are repeated from our previous publication \citep{prachaseree2022learning}. Across all datasets shown in Fig. \ref{fig:calibration}a-c, direct comparison of the I and E-M1 points convincingly indicates that ensemble averaging holds up as a strategy for improving model performance and calibration. {\color{revs}As outlined in section \ref{sec:methods_ensemble}, we would like to emphasize that during the training of each individual network, the weights were randomly initialized. The impact of this random initialization can be observed in figure \ref{fig:calibration}. However, it is worth noting that the use of ensemble averaging has the potential to mitigate the effects of such randomness and effectively nullify any potential impact of random weight initialization.}
These positive results, all without post hoc calibration, serve as a baseline for evaluating the efficacy of methods designed specifically to improve calibration, discussed in the next Section. 

\subsubsection{Temperature Scaling After Ensemble Averaging Offers Limited Improvement to Deep Learning Model Calibration}

In Fig. \ref{fig:calibration}, we also show the results of three approaches to performing post hoc model calibration via temperature scaling (I-C, E-M2, and E-M3), where the individual models (I) and models with straightforward ensemble averaging (E-M1) serve as the baseline. From Fig. \ref{fig:calibration}, we see that unlike the results of applying ensemble averaging, the results of applying temperature scaling are much less consistent. {\color{revs-minor} In contrast with machine learning literature \citep{guo2017calibration},} when applying temperature scaling to individual neural networks (i.e., comparing the I models to the I-C models), there were limited and inconsistent benefits. For example, in Fig. \ref{fig:calibration}b, temperature scaling leads to limited decreases in ECE for individual networks, while in Fig. \ref{fig:calibration}c temperature scaling leads to a modest increase in ECE. 
And, in comparing individual models with temperature scaling (I-C) to ensemble models without temperature scaling (E-M1), we note that the E-M1 models more consistently lead to both lower ECE and lower error.

When comparing ensemble averaging combined with post hoc calibration via temperature scaling (E-M2 and E-M3) to the baseline of straightforward ensemble averaging (E-M1), we find similar inconsistent results. 
Specifically, ensemble averaging combined with temperature scaling (E-M2, E-M3) lead to inconsistent performance improvements in comparison to ensemble averaging alone (E-M1). 
And, temperature scaling prior to ensemble averaging (E-M2) led to strikingly worse performance compared to ensemble averaging alone (E-M1) for the Mechanical MNIST -- Crack Path dataset, illustrated in Fig. \ref{fig:calibration}c. Notably, this dramatic difference may be due to the severe class imbalance present in the Mechanical MNIST -- Crack Path dataset. 
Overall, we found that for these datasets ensemble averaging offers much more consistent performance improvements than post hoc model calibration via temperature scaling.
The supplementary reliability diagrams shown in Appendix \ref{sec:apx_1}, Fig. \ref{fig:reliability_2} also support these results. In addition, it is worth mentioning that results shown in Fig. \ref{fig:training-size}, where increasing the training set size led to improvements in both model error and model ECE for ensemble neural networks, also indicate that increasing the initial training set size may be a better use of data resources than post hoc calibration via temperature scaling. However, we acknowledge that this statement may vary based on specific desired outcomes and data resources.

Overall, we note that our findings are based on \emph{empirical} evidence obtained through analyzing the $7$ datasets introduced in this manuscript. Broadly speaking, the trends observed in this study are not necessarily guaranteed to be consistent with either previous or forthcoming literature that is based on different data. For example, others have empirically shown that temperature scaling applied after ensemble averaging (M3) can consistently improve model calibration compared to ensemble averaging alone (M1) \citep{rahaman2020uncertainty}. {\color{revs-minor} However, there is no real consensus on the best method to calibrate deep learning models, and calibration strategies to date appear dependent on the type of dataset and deep learning architecture used \citep{guo2017calibration, lakshminarayanan2017simple, minderer2021revisiting,ovadia2019can, rahaman2020uncertainty, zhang2020mix}. As such, } 
rather than taking the results of this investigation at face value, we hope that our work (1) highlights the need for more research towards understanding deep neural network calibration, and (2) emphasizes the need for additional domain specific open access datasets for systematically exploring the efficacy of deep learning approaches.

\section{Conclusion}
\label{sec:conclusion} 

To the author's knowledge, this is the largest investigation to date of deep learning model calibration for classification problems in mechanics. From this investigation, we found four key results. First, we found that ensemble neural networks perform comparably to Gaussian Process Classification and Support Vector Classification with Platt scaling in terms of model error and model expected calibration error (ECE) for all three BIC datasets. Second, we found that increasing the training set size decreases both model error and model ECE for all three BIC datasets. Third, we found that ensemble averaging consistently improves both model error and model ECE for all $7$ datasets. Fourth, we found that temperature scaling offers limited benefits in comparison to ensemble averaging for all $7$ datasets. Overall, we believe that this work demonstrates the utility of large scale studies of {\color{revs-minor}machine learning} methods applied to problems in mechanics.

Looking forward, we anticipate {\color{revs-minor}several} major areas of future investigation by both us and others. First, these datasets can be used to investigate alternative approaches to simultaneously improving both model error and ECE. For example, there are multiple approaches to post hoc model calibration beyond temperature scaling that are amenable to similar investigation \citep{kuleshov2018accurate, naeini2015obtaining, rahimi2020post}. Alternatively, Bayesian methods \citep{kendall2017uncertainties, maddox2019simple, zhang2021bayesian} and evidential deep learning models \citep{amini2020deep, sensoy2018evidential} aim to output calibrated predictions without any additional post hoc training. And, building on exciting recent work \citep{raissi2019physics, yang2019adversarial}, we anticipate that there are rich possibilities for physics-informed approaches to this problem. Second, these datasets can be used to investigate alternative approaches to evaluating model calibration. As stated in Section \ref{sec:ece_note}, developing more effective metrics remains an open area of research and one that deserves attention in a mechanics-specific context. Third, there is a need to extend this study to additional open-access mechanics-based datasets from diverse sources. {\color{revs}Ultimately, we acknowledge that the current study is limited to data generated through FEA, which introduces a potential bias. This highlights the need for further research to explore the problem with data from other sources, such as experimental testing or molecular dynamic simulations, which may contain stochastic behavior. Additionally, our study is limited to simulated data since, to our knowledge, there are currently no open access experimental mechanics datasets that are both readily accessible and sufficiently large to include in this study.} Overall, our hope is that this work both offers a starting point for researchers beginning work with deep learning model calibration, and motivates future mechanics-specific advances in deep learning model calibration. Because all datasets and codes associated with this manuscript are available under open-source licenses, others can readily build on our work and make direct comparisons to alternative methods.

\section{Additional Information}
\label{sec:additional} 

{\color{revs}
All datasets used in this investigation have been previously published in conjunction with prior manuscripts from our group \citep{lejeune2021geometric,prachaseree2022asymmetric,mohammadzadeh2022predicting}. Each dataset contains both the metadata to interpret files and the code needed to reproduce all results. In all cases, data is shared under a CC BY-SA 4.0 License through the OpenBU Institutional Repository and code is shared under a MIT License through GitHub. The datasets used are as follows: 
\begin{itemize}
    \item \textbf{Buckling Instability Classification (BIC)} \citep{lejeune2020buckling}: rectangular columns with heterogeneous material properties are subject to a fixed level of applied displacement and classified as either stable or unstable (i.e., buckled).
    
        BIC contains three \textit{independent} sub-datasets where all three $16 \times 1$ input patterns are sampled from different distributions:
        \begin{itemize}
            \item[] \textbf{BIC 1}: $16 \times 1$ pattern from distribution 1, sampling 2 discrete values (input) $\mapsto$ stable vs. unstable (output)
            \item[] \textbf{BIC 2}: $16 \times 1$ pattern from distribution 2, sampling 3 discrete values (input) $\mapsto$ stable vs. unstable (output)
            \item[] \textbf{BIC 3}: $16 \times 1$ pattern from distribution 3, sampling continuous range  (input) $\mapsto$ stable vs. unstable (output)
        \end{itemize}
    \item \textbf{Asymmetric Buckling Columns (ABC)} \citep{prachaseree2022asymmetric}: heterogeneously architected and asymmetric columns with homogeneous material properties are subject to a fixed level of applied displacement and classified as either left buckling or right buckling. 

    ABC contains three \textit{independent} sub-datasets where all three input domain architecture types are generated through different procedural approaches:
        \begin{itemize}
            \item[] \textbf{ABC 1}: spatial graph that represents domains from domain type 1, block stacking (input) $\mapsto$ left vs. right (output)
            \item[] \textbf{ABC 2}: spatial graph that represents domains from domain type 2, uniform rings (input) $\mapsto$ left vs. right (output)
            \item[] \textbf{ABC 3}: spatial graph that represents domains from domain type 3, clipped non-uniform rings (input) $\mapsto$ left vs. right (output)
        \end{itemize}
    \item \textbf{Mechanical MNIST -- Crack Path} \citep{mohammadzadeh2021mechanical}: two-dimensional square domains with heterogeneous material properties and a defined initial crack are subject to a fixed level of applied displacement. Under these loading conditions, a crack propagates throughout the domain with the crack path dictated by the heterogenous material property distribution. 

    Mechanical MNIST -- Crack Path is a single dataset:
        \begin{itemize}
            \item[] \textbf{Mechanical MNIST -- Crack Path}: $64 \times 64$ material property array (input) $\mapsto$ $64 \times 64$ damage field (output)
        \end{itemize}
\end{itemize}
All datasets are schematically illustrated in Fig. \ref{fig:datasets} for a total of $7$ independently trained and tested cases \citep{lejeune2020buckling,prachaseree2022asymmetric,mohammadzadeh2021mechanical}.

The code to reproduce the novel computational results presented in this paper is available through GitHub (\url{https://github.com/saeedmhz/model-calibration}) shared under a MIT License.} 

\section{Acknowledgements}
\label{sec:ack} 
We would like to thank the staff of the Boston University Research Computing Services and the OpenBU Institutional Repository (in particular Eleni Castro) for their invaluable assistance with generating and disseminating the datasets used in this paper. This work was made possible through start up funds from the Boston University Department of Mechanical Engineering, the David R. Dalton Career Development Professorship, the Hariri Institute Junior Faculty Fellowship, the Haythornthwaite Research Initiation Grant, the National Science Foundation Grant CMMI-2127864, the American Heart Association Career Development Award 856354, and the Office of Naval Research Grant N00014-22-1-2066.

\appendix
{\color{revs}
\section{List of Abbreviations}
\begin{itemize}
    \item[] \textbf{FEA} - Finite Element Analysis
    \item[] \textbf{BIC} - Buckling Instability Classification
    \item[] \textbf{ABC} - Asymmetric Buckling Columns
    \item[] \textbf{NLL} - Negative Log Likelihood
    \item[] \textbf{ECE} - Expected Calibration Error
    \item[] \textbf{GPC} - Gaussian Process Classification
    \item[] \textbf{SVC} - Support Vector Classification
    \item[] \textbf{NN-I} - Individual neural network (see figure \ref{fig:bic-err-vs-ece} and \ref{fig:training-size})
    \item[] \textbf{NN-E} - Ensemble neural network (see figure \ref{fig:bic-err-vs-ece} and \ref{fig:training-size})
    \item[] \textbf{Method (I)} - Individual neural network without post hoc calibration (see figure \ref{fig:calibration})
    \item[] \textbf{Method (I-C)} - Individual neural network with post hoc calibration via temperature scaling (see figure \ref{fig:calibration})
    \item[] \textbf{Method (E-M1)} - Ensemble neural network without post hoc calibration (see figure \ref{fig:calibration})
    \item[] \textbf{Method (E-M2)} - Ensemble neural network with post hoc calibration via temperature scaling applied before
ensemble averaging (see figure \ref{fig:calibration})
    \item[] \textbf{Method (E-M3)} - Ensemble neural network with post hoc calibration via temperature scaling applied after en-
semble averaging (see figure \ref{fig:calibration})
\end{itemize}
\label{sec:apx_1}
}
\section{Supplementary Reliability Diagrams}
\label{sec:apx_1} 
\begin{figure}[p]
\centering
\includegraphics[width=0.995\textwidth,height=\textheight,keepaspectratio]{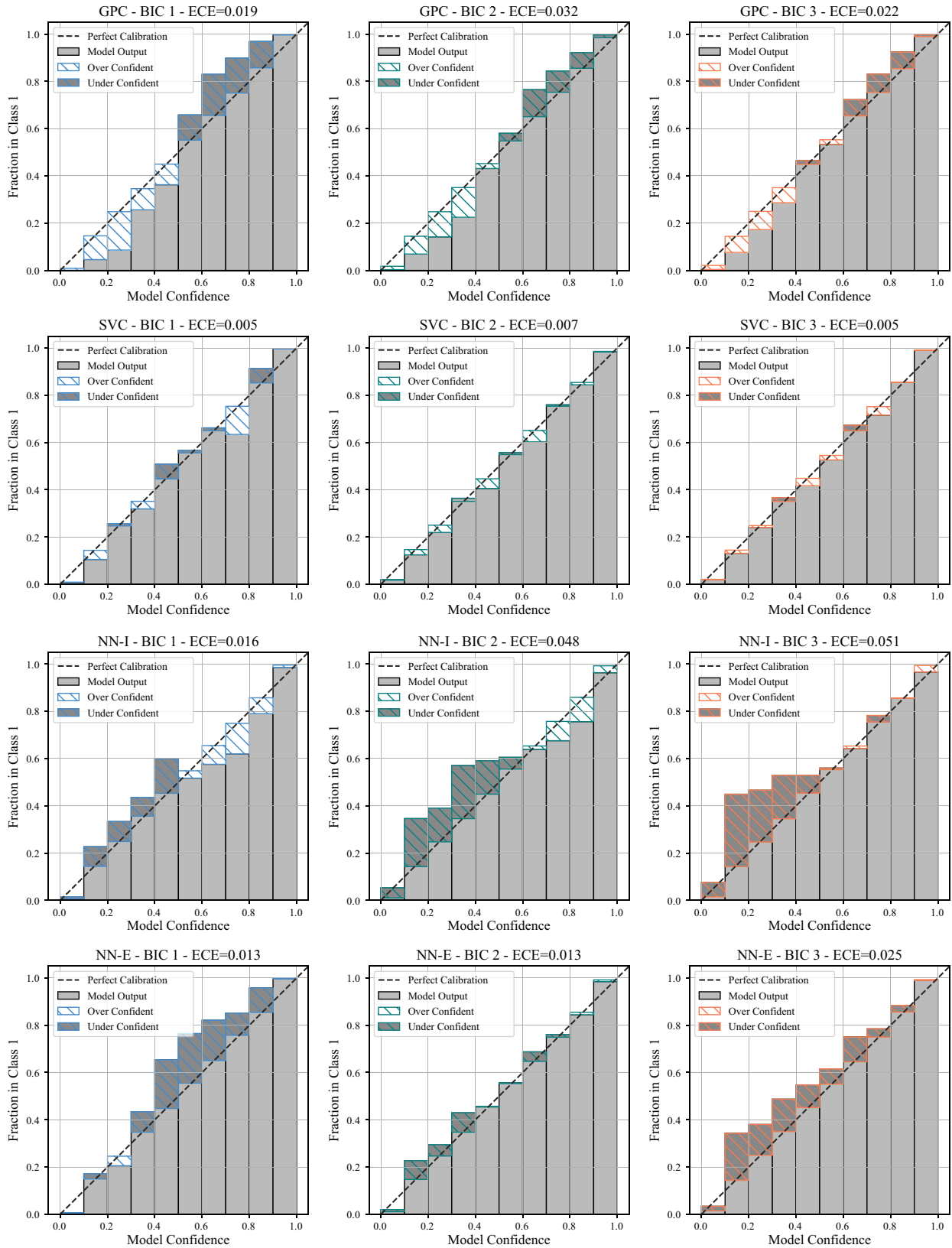}
\caption{Reliability diagrams as a supplement to Fig. \ref{fig:bic-err-vs-ece}.}
\label{fig:reliability_1}
\end{figure}

\begin{figure}[p]
\centering
\includegraphics[width=0.995\textwidth,height=\textheight,keepaspectratio]{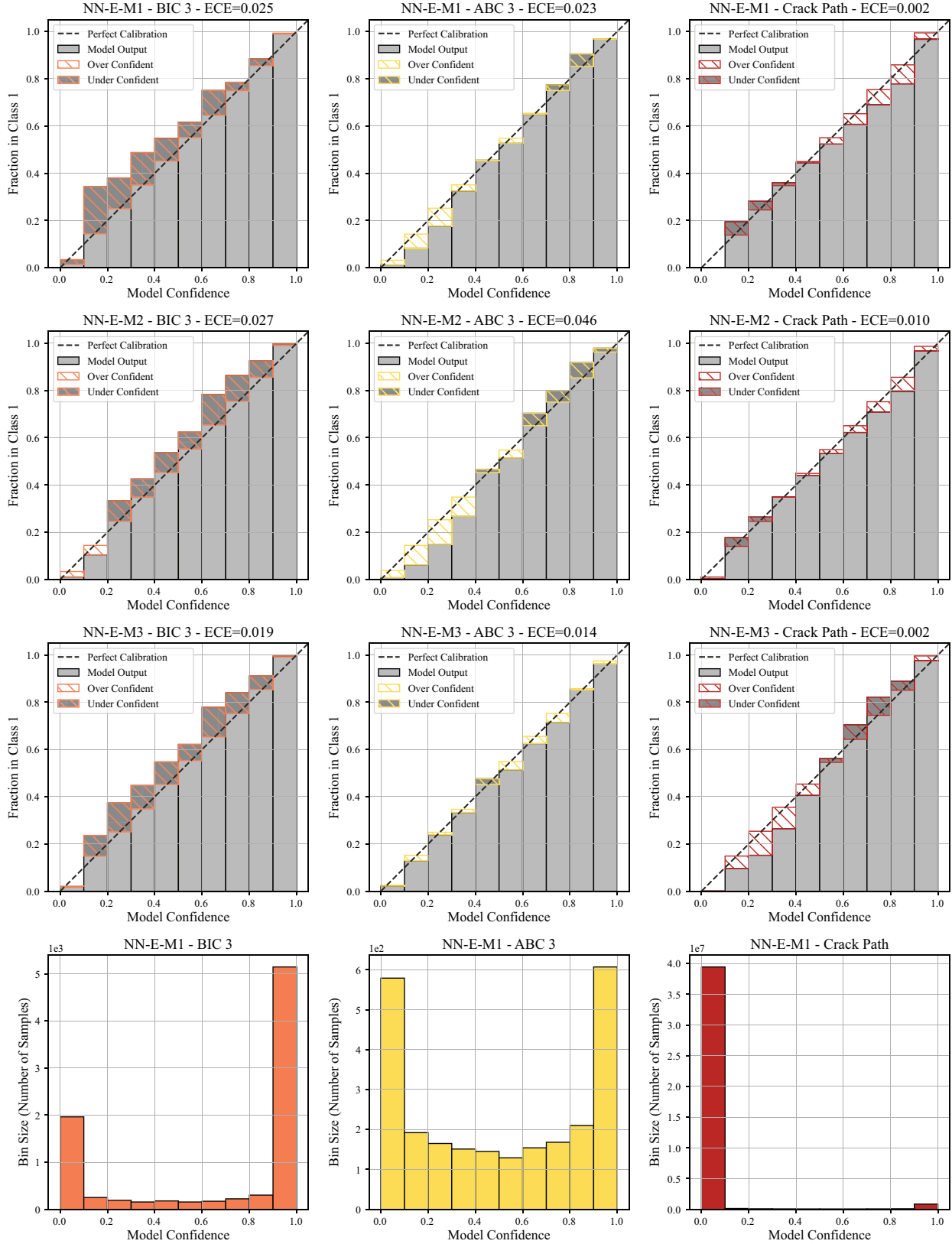}
\caption{Reliability diagrams and confidence distribution histograms as a supplement to Fig. \ref{fig:calibration}.}
\label{fig:reliability_2}
\end{figure}

In Section \ref{sec:results}, we present the core results of this investigation as plots of Error vs. Expected Calibration Error (ECE). As introduced in Section \ref{sec:ece} and illustrated in Fig. \ref{fig:reliability}, ECE is connected to the reliability diagram, a common strategy for visualizing model calibration. Here we provide supplementary reliability diagrams to accompany Fig. \ref{fig:bic-err-vs-ece} and Fig. \ref{fig:calibration}. In Fig. \ref{fig:reliability_1}, we show $12$ calibration curves that correspond to the results shown in Fig. \ref{fig:bic-err-vs-ece}. Specifically, for BIC 1, BIC 2, and BIC 3 we plot reliability diagrams for Gaussian Process Classification, Support Vector Classification with Platt scaling, a representative Individual Neural Network, and an Ensemble Neural Network. In Fig. \ref{fig:reliability_2}, we show $9$ calibration curves that correspond to the results shown in Fig. \ref{fig:calibration}. Specifically, for BIC 3, ABC 3, and Mechanical MNIST Crack Path we plot reliability diagrams for straightforward ensemble averaging (E-M1), post hoc calibration temperature scaling followed by ensemble averaging (E-M2), and ensemble averaging followed by post hoc calibration temperature scaling (E-M3). In addition, we plot histograms of the representative distribution of model confidence for E-M1 for BIC 3, ABC 3, and Mechanical MNIST Crack Path. These histograms not only indicate the bins that will have the heaviest weight, but also illustrate the severe class imbalance present in the Mechanical MNIST Crack Path dataset.

Beyond overall ECE, which is a weighted average of the calibration error in each bin, the reliability diagram allows us to visualize the Maximum Calibration Error and the regions where the different models tend to be over and under confident. In addition, they provide a visualization of the outcomes of post hoc model calibration via temperature scaling. These diagrams provide additional contextual information that helps address some of the limitations of ECE as a metric, discussed in Section \ref{sec:ece_note}. Overall, we recommend visualizing the reliability diagram in addition to computing ECE prior to deploying a given model.

{\color{revs}
\section{Details of the Neural Networks}
\label{sec:apx_neural_nets} 
To supplement the description of the neural networks we used in this work, introduced in Sections \ref{sec:fnn}, \ref{sec:gnn}, and \ref{sec:unet}, we have included network schematics for our implementations of each network in Fig. \ref{fig:app-networks}. 

\begin{figure}[h]
\centering
\includegraphics[width=0.85\textwidth, height=0.85\textheight, keepaspectratio]{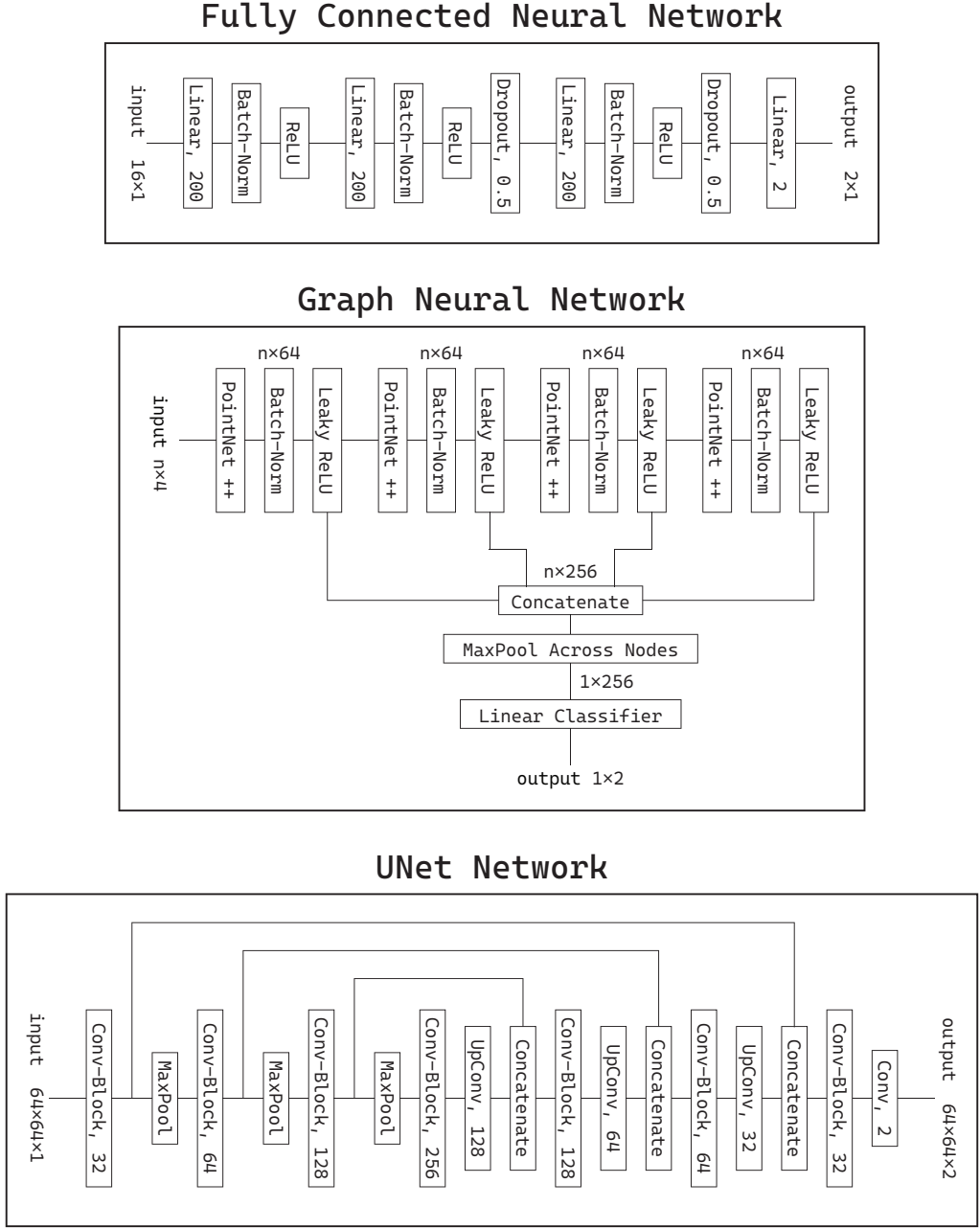}
\caption{{\color{revs}This figure supplements Sections \ref{sec:fnn}, \ref{sec:gnn}, and \ref{sec:unet} by schematically illustrating the different neural network architectures. The fully connected neural networks consists of three linear layers, each with $200$ nodes, and utilizes dropout with a probability of $0.5$ during training. Our previous work \citep{prachaseree2022learning} provides further details on the graph neural network that we used. Fiinally, the UNet network employs ``Conv-blocks,'' which repeat a 2D convolutional layer followed by a 2D batch normalization and a ReLU activation function twice. The convolutional layers have a filter size of $3$ and a stride and padding of $1$, while the maxpool layers use a filter size of $2$. The transposed convolutional (``UpConv'') layers employ a filter and stride of $2$.}}
\label{fig:app-networks}
\end{figure}
}
\clearpage 
\newpage

\bibliographystyle{plain}
\bibliography{main}
   
\end{document}